\documentclass[11pt,a4paper]{article}

\usepackage[T1]{fontenc}
\usepackage[utf8]{inputenc}
\usepackage{lmodern}
\usepackage{microtype}
\usepackage[margin=2.5cm]{geometry}
\usepackage{amsmath,amssymb}
\usepackage{xcolor}
\usepackage{booktabs}
\usepackage{array}
\usepackage{tabularx}
\usepackage{enumitem}
\usepackage{caption}
\usepackage{tikz}
\usetikzlibrary{positioning,arrows.meta,shapes.geometric,fit,backgrounds,calc}
\usepackage[numbers,square,sort&compress]{natbib}
\usepackage[hidelinks]{hyperref}
\usepackage[capitalise,noabbrev]{cleveref}
\usepackage{authblk}

\setlist{itemsep=1pt,topsep=3pt,parsep=0pt}
\newcolumntype{L}[1]{>{\raggedright\arraybackslash}p{#1}}
\newcolumntype{Y}{>{\raggedright\arraybackslash}X}
\crefname{figure}{figure}{figures}
\Crefname{figure}{Figure}{Figures}
\crefname{table}{table}{tables}
\Crefname{table}{Table}{Tables}
\crefname{section}{section}{sections}
\Crefname{section}{Section}{Sections}

\definecolor{cblue}{HTML}{2A78D6}
\definecolor{corange}{HTML}{EB6834}
\definecolor{caqua}{HTML}{1BAF7A}
\definecolor{cgood}{HTML}{0CA30C}
\definecolor{cwarn}{HTML}{FAB219}
\definecolor{cserious}{HTML}{EC835A}
\definecolor{ccrit}{HTML}{D03B3B}
\definecolor{cgray}{HTML}{6B6B6B}
\definecolor{clight}{HTML}{EEF4FC}
\definecolor{cminor}{HTML}{9EC5F4}

\newcommand{\verdict}[1]{\textsf{\small #1}}
\tikzset{every picture/.append style={execute at begin picture={\sffamily}}}

\title{Trust, but verify, audit and attribute: a human-in-the-loop framework for checking AI-produced research results}

\author[1,2]{Gy\"orgy Eigner\thanks{Corresponding author. E-mail: \href{mailto:eigner.gyorgy@uni-obuda.hu}{eigner.gyorgy@uni-obuda.hu}}}
\author[1,3]{Patrik P. S\"uli}
\author[4]{Roland Holl\'os}
\affil[1]{Biomatics and Applied Artificial Intelligence Institute, John von Neumann Faculty of Informatics, Obuda University, Budapest, Hungary}
\affil[2]{Physiological Controls Research Center, Obuda University, Budapest, Hungary}
\affil[3]{Doctoral School of Applied Informatics and Applied Mathematics, Obuda University, Budapest, Hungary}
\affil[4]{HUN-REN Centre for Agricultural Research, Brunszvik u.~2., 2462 Martonv\'as\'ar, Hungary}
\date{}

\begin{document}
\maketitle

\begin{abstract}
\noindent
Artificial intelligence (AI) systems now produce research results in bulk: in October 2026 a single laboratory released more than 700 AI-generated mathematics manuscripts, only part of them with machine-checked proofs. Such volumes will be checked mostly by AI agents that call specialised software. A passing check certifies only the artefact that was checked. For proofs translated into the proof assistant Lean, it has been shown that this artefact may state something other than the author's text, and that deciding whether a translation is faithful is undecidable in general. The human researcher therefore stays in the loop, and human and AI contributions must be recorded. We propose a domain-independent, human-in-the-loop framework for checking AI-produced results. Validated, agent-usable tools with different grounds of trust check the results; a fidelity audit asks whether each check addressed the claim as stated; the human decides at fixed points; and a dialog report attributes the work to the human and to the AI. We describe two automatic implementations of the tool layer, for mathematics (Wolfram Language, MATLAB, Lean) and statistics (Python and R), and run the mathematical one on a two-claim example. An ambiguous claim was split into two readings, one refuted and one proved.

\medskip\noindent\textbf{Keywords:} AI-generated research, verification, semantic fidelity, human-in-the-loop, autoformalisation, agent tools, human--AI attribution
\end{abstract}

\section{Introduction}\label{sec:intro}

Large language models (LLMs) and the artificial intelligence (AI) agents built on them now derive results, write code, and run analyses for research papers. In recent times, the scale has changed and agentic scientific work has sped up. In October 2026 OpenAI released, in one public repository, 719 AI-generated mathematics manuscripts in 372 families \cite{openai2026math,nature2026openai,zrubka2023artificial}. According to the repository, about 42\% of the top-line results have Lean formalisations, and ``some of the unformalized results could have issues'' \cite{openai2026math}. Such a release exceeds what human referees can read in any reasonable time. If AI-produced results arrive at this rate, most of them will be checked, if at all, by software, and increasingly by AI agents that call that software.

Whether such checks mean anything is, at its core, a question of translation between what the user wants the LLM to do and what really happens. A proof assistant that accepts a proof, a statistics package that returns a test result and a simulator that returns a converged solution all give a strong signal. But each tool checks only what it was given, and what it was given was written by the AI from the researcher's text. The check is therefore only as good as the answer to one question: did the AI translate correctly what the researcher wanted, for example the theorem into a formal statement, the analysis plan into a script or the model into a simulation setup? Software engineering knows the same problem. Code generated by an LLM can pass the tests it was shown without implementing the requirement. When the test suites of a standard benchmark were made more thorough, a substantial share of solutions that had counted as correct failed \cite{liu2023evalplus}. In research, the requirement is a claim or hypothesis or thesis in a paper, and there is usually no test suite at all.

Bastounis et al.~\cite{bastounis2026} make this precise for formal mathematics. Finding a Lean proof for a correctly stated theorem is one task; translating a mathematical text into Lean so that its meaning is preserved is another, and success in the first says nothing about the second. The AI may prove a slightly different statement, something similar to what the researcher wants, but not exactly what the researcher intended, or the right statement by a different argument, and Lean accepts both. The authors document such mismatches in a published Lean formalisation of a claimed Navier--Stokes blow-up proof, and they prove that resolving ambiguities in a natural-language definition is, in general, harder than the Halting problem. No AI translator can be faithful on every input, and no algorithm can always tell when a translation is unfaithful.

Two consequences shape this paper. First, the human cannot be fully removed from forming and checking scientific results. Where fidelity cannot be decided by an algorithm, a person responsible for the claim has to decide which reading is meant and whether a check addressed it. Second, once a human and an AI produce and check a result together, it must be known which parts are human work and which are AI work. This is especially important because an LLM synthesises knowledge from existing work differently from a researcher, and LLM-based systems can mislead knowledge generation: they make microdecisions during the research flow, and these decisions can lead away from what the researcher wanted to achieve in the assisted research session. Journals already require a statement of AI use \cite{nature2023rules}, and questions of credit arise as soon as AI output is published in quantity \cite{nature2026openai}. A framework for checking AI-produced results must therefore check the results with reliable tools, make the remaining fidelity questions visible to a human, and record who contributed what.

We propose a framework with a human in the loop for checking AI-produced research results (\cref{sec:pipeline}). It has four parts:
\begin{enumerate}[label=(\roman*), leftmargin=*, itemsep=1pt]
\item validated, agent-usable tools that rest on different grounds of trust check the results;
\item a fidelity audit asks, claim by claim, whether each check verified what the author stated (\cref{sec:audit});
\item human verification at four fixed decision points settles what no algorithm can (\cref{sec:human});
\item a human--AI dialog report records which decisions came from the human and which from the AI (\cref{sec:dialog}).
\end{enumerate}
The framework is a methodology, not a particular toolchain: it does not depend on the domain, and it can be instantiated in any field where validated tools exist that an agent can call. We show how this is done with two examples built according to the methodology, \texttt{math-prove-tool} for mathematics (Wolfram Language, MATLAB, Lean) and \texttt{stat-prove-tool} for statistics (a validated Python server and R server) (\cref{sec:toolchains}), and we follow one small run of the first through the framework (\cref{sec:example}). We do not claim that the framework guarantees fidelity, which Bastounis et al.~\cite{bastounis2026} rule out; we claim that it makes the remaining risk explicit and leaves its resolution to a human.

\section{The problem: a passing check is not a faithful check}\label{sec:problem}

\subsection{Claim, artefact and verdict}
Every verification step involves three objects (\cref{fig:gap}), whatever the domain. The \emph{claim} is what the author states in the paper, in words, formulas, tables and figures. The \emph{artefact} is what is run: a formal theorem, a symbolic computation, a statistical script or a simulation model. The \emph{verdict} is what the tool reports: ``proof accepted'', ``\texttt{True}'', ``$p<0.05$'' or ``converged''.

A validated tool's verdict is reliable for the second gap, between artefact and verdict: if a proof assistant accepts a proof, the formal statement follows from the axioms. No tool says anything about the first gap, between claim and artefact. If the artefact states a different theorem, uses a different subset of the data or checks only an instance, the verdict is still correct, but about something else.

\begin{figure}[t]
\centering
\begin{tikzpicture}[
  box/.style={draw, rounded corners=2pt, align=center, font=\footnotesize, text width=28mm, minimum height=15mm},
  arr/.style={-{Stealth[length=2mm]}, line width=0.8pt}
]
\node[box, draw=cblue, line width=0.9pt] (c) {\textbf{Claim}\\as the author states it\\(paper)};
\node[box, draw=corange, line width=0.9pt, right=18mm of c] (a) {\textbf{Artefact}\\formal theorem, script,\\model, analysis};
\node[box, draw=cgood, line width=0.9pt, right=18mm of a] (v) {\textbf{Verdict}\\``accepted'', \texttt{True},\\``$p<0.05$'', ``converged''};
\draw[arr, dashed, draw=ccrit] (c) -- node[above, font=\footnotesize, text=ccrit, align=center]{gap 1:\\fidelity} node[below, font=\footnotesize, text=ccrit, align=center]{no tool\\checks this} (a);
\draw[arr, draw=cgood] (a) -- node[above, font=\footnotesize, text=cgood!70!black, align=center]{gap 2:\\soundness} node[below, font=\footnotesize, text=cgood!70!black, align=center]{the tool\\checks this} (v);
\node[font=\footnotesize, align=left, text width=120mm, below=8mm of a] (ex) {%
Ways in which gap~1 opens. Documented for proofs \cite{bastounis2026}:\\[1pt]
$\bullet$ different statement: $m+5$ instead of $m+4$ derivatives in a Lean estimate\\
$\bullet$ hidden default: $\inf\emptyset=0$ in Lean makes an undefined object look defined\\
$\bullet$ one-word change: ``odd'' read as ``prime'', and the formal statement still compiles\\[2pt]
\textit{The same pattern in other settings (illustrations):}\\
$\bullet$ statistics: another test, sample, exclusion rule or correction than the paper states\\
$\bullet$ code: passes the given tests but does not implement the requirement \cite{liu2023evalplus}};
\end{tikzpicture}
\caption{The two gaps in every verification step. A validated tool closes the gap between the artefact and its verdict (solid arrow). Nothing in the tool closes the gap between the author's claim and the artefact (dashed arrow); this is the gap the fidelity audit examines.}
\label{fig:gap}
\end{figure}

Why does gap~1 open? When success is judged only by whether Lean accepts the result, the AI keeps revising its translation until it compiles, and this loop rewards replacing the author's argument as much as translating it \cite{bastounis2026}. A second cause is plausible but not established here: LLMs tend to produce what the user appears to want; owing to the architecture of leading LLMs, the user's wishes are probability-based predictions \cite{sharma2024sycophancy}. Scale makes both worse: a 2026 project by Rammal et al.\ that translated 26 textbooks into Lean, with semantic checking done mainly by other AIs, was found to contain many incorrectly formalised statements, as reported in~\cite{bastounis2026}.

\subsection{Fidelity}
We call an artefact \emph{statement-faithful} to a claim if it establishes the claim with the same objects, the same quantifiers in the same order, the same hypotheses, the same conclusion including every quantitative detail, and the same presuppositions (every object the claim names exists and is well defined). In statistics the quantitative detail includes the population, the test, the error rate and the multiple-comparison correction; in simulation, the equations, parameters, boundary conditions and discretisation. We call an artefact \emph{argument-faithful} if, in addition, it follows the author's route: the author's statements, ideas, proof, analysis plan or modelling choices. An author who wants to know whether a result is true needs statement fidelity; an author who wants to know whether the derivation they wrote is right needs both.

These definitions say what must be compared: they cannot always be decided, especially in formal mathematics \cite{bastounis2026}. Three consequences follow, and the framework is built around them.
\begin{enumerate}[label=\arabic*., leftmargin=*]
\item A successful tool run closes gap~2 only. A separate step must ask, claim by claim, whether gap~1 is closed.
\item That step cannot be fully automatic. Many fidelity questions have an answer only after a human chooses between two readings of an ambiguous sentence, for example whether ``the largest value is $0.37$'' means the exact value or a rounded one.
\item Agreement between AI systems, for example a solution based on a council of AI models, or between tools with the same ground of trust is not evidence of fidelity. Two systems that read the claim the same wrong way will agree. Agreement between tools counts as evidence about gap~2 only when the tools rest on different grounds of trust, and never as evidence about gap~1.
\end{enumerate}

\section{The framework}\label{sec:pipeline}

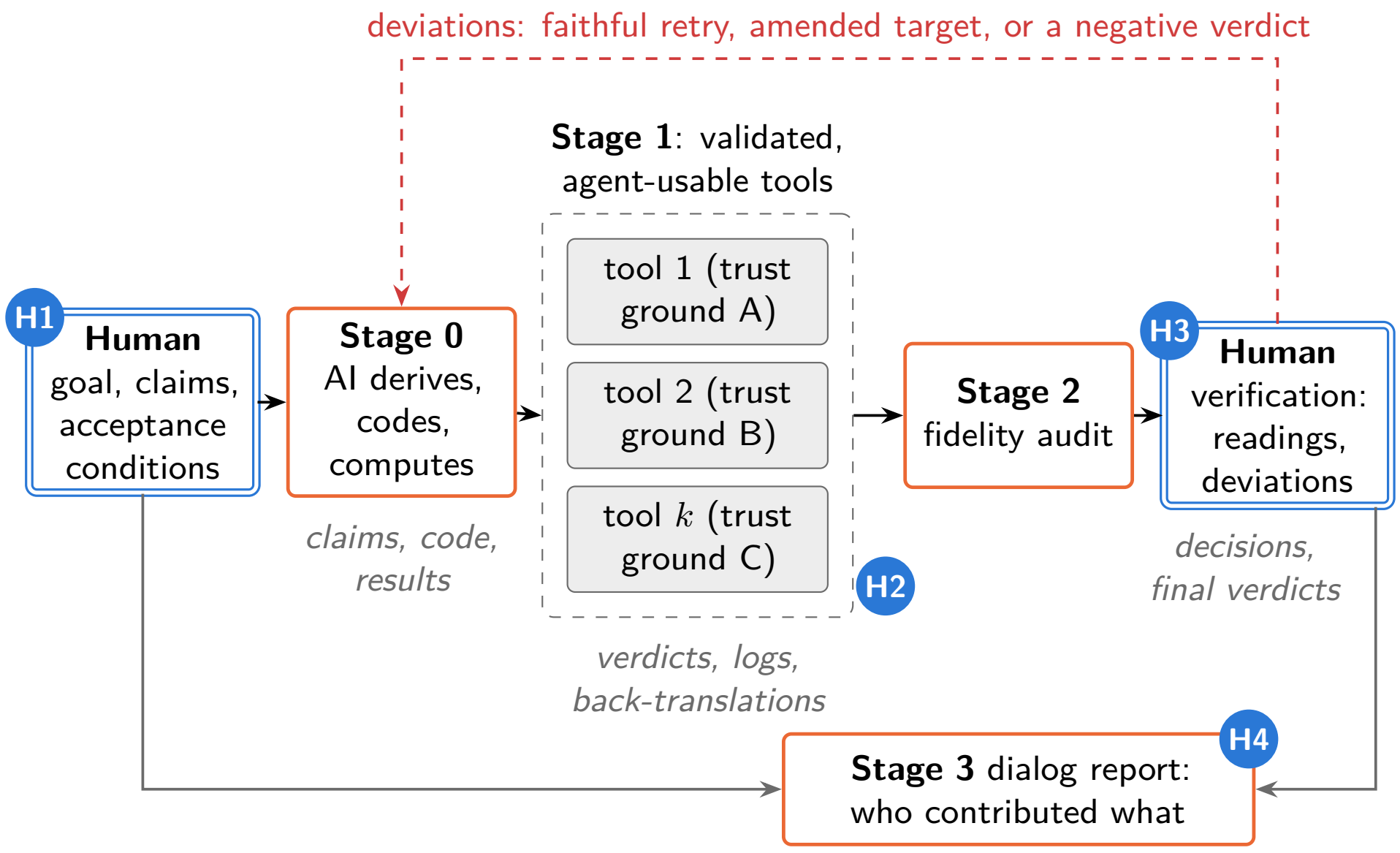
\begin{figure}[t]
\centering
\begin{tikzpicture}[
  node distance=5mm and 2.6mm,
  stage/.style={draw, rounded corners=2pt, align=center, font=\footnotesize, minimum height=13mm, text width=18mm, fill=white},
  human/.style={stage, draw=cblue, line width=0.6pt, double, double distance=1pt},
  ai/.style={stage, draw=corange, line width=0.9pt},
  tool/.style={stage, draw=cgray, fill=cgray!12, text width=21mm, minimum height=7mm, font=\footnotesize},
  hp/.style={circle, draw=cblue, fill=cblue, text=white, font=\scriptsize\bfseries, inner sep=1pt, minimum size=4.6mm},
  outlbl/.style={font=\footnotesize\itshape, text=cgray, align=center},
  arr/.style={-{Stealth[length=2mm]}, line width=0.7pt},
  back/.style={-{Stealth[length=2mm]}, line width=0.7pt, dashed, draw=ccrit}
]
\node[human] (h) {\textbf{Human}\\goal, claims,\\acceptance\\conditions};
\node[ai, right=of h] (s0) {\textbf{Stage 0}\\AI derives,\\codes, computes};
\node[tool, right=4.5mm of s0, yshift=10mm] (w) {tool 1 (trust ground A)};
\node[tool, below=1.5mm of w] (m) {tool 2 (trust ground B)};
\node[tool, below=1.5mm of m] (l) {tool $k$ (trust ground C)};
\begin{scope}[on background layer]
\node[draw=cgray, dashed, rounded corners=3pt, fit=(w)(m)(l), inner sep=2.2mm, label={[font=\footnotesize, align=center]above:\textbf{Stage 1}: validated,\\agent-usable tools}] (s1) {};
\end{scope}
\node[ai, right=4.5mm of s1] (s2) {\textbf{Stage 2}\\fidelity audit};
\node[human, right=of s2] (d) {\textbf{Human}\\verification:\\readings,\\deviations};
\node[ai, below=22mm of s2, text width=40mm, minimum height=10mm] (s3) {\textbf{Stage 3} dialog report:\\who contributed what};
\draw[arr] (h) -- (s0);
\draw[arr] (s0) -- (s1.west);
\draw[arr] (s1.east) -- (s2);
\draw[arr] (s2) -- (d);
\draw[back] (d.north) -- ++(0,24mm) -| node[pos=0.25, above, font=\footnotesize, text=ccrit]{deviations: faithful retry, amended target, or a negative verdict} (s0.north);
\draw[arr, draw=cgray] ([xshift=-1.5mm]d.south east) |- (s3.east);
\draw[arr, draw=cgray] (h.south) |- (s3.west);
\node[hp, above left=-2.5mm and -2.5mm of h] {H1};
\node[hp, right=1mm of s1.south east, anchor=south west, yshift=1mm] {H2};
\node[hp, above left=-2.5mm and -2.5mm of d] {H3};
\node[hp, above right=-2.5mm and -2.5mm of s3] {H4};
\node[outlbl, below=1mm of s0] {claims, code,\\results};
\node[outlbl, below=1mm of s1] {verdicts, logs,\\back-translations};
\node[outlbl, below=1mm of d, xshift=-3mm] {decisions,\\final verdicts};
\end{tikzpicture}
\caption{The framework. Double frames (blue): human roles; single frames (orange): AI roles; shaded frames (grey): Stage-1 tools, chosen to rest on different grounds of trust. Filled circles H1--H4: the four human decision points of \cref{sec:human}. Dashed arrow: deviations returned for a faithful retry, an amended target or a negative verdict. Arrows into Stage~3: records that feed the dialog report.}
\label{fig:pipeline}
\end{figure}

\Cref{fig:pipeline} shows the framework. We walk through it from left to right and then state which tools may enter the tool layer.

\subsection{Roles}
The framework has four roles. The \emph{human} owns the research domain and questions, the claims, the acceptance conditions and every decision that changes what is claimed, which is why it is critical to assess which decisions were made by the human researcher and which by the AI; the human does not need to write code or proofs. The \emph{AI} derives, codes, computes, writes the verification artefacts and drafts the reports; its output is a proposal until the later stages have examined it. The \emph{tools} check artefacts, and their verdicts concern the artefact only. The \emph{auditor} examines whether each artefact addresses the claim as stated; it may be an AI working with restricted inputs, but it does not take the human's decisions.

\subsection{The four stages}
\begin{enumerate}[label=Stage~\arabic*., start=0, leftmargin=*, itemsep=1pt]
\item The human states the goal, the claims and the acceptance conditions (decision point H1), and the AI produces the result. Nothing produced here is trusted on its own. When AI-generated results arrive in bulk, this stage has happened elsewhere, and the framework starts from the published claims.
\item Several validated, agent-usable tools check the claims under three rules. The list of claims is frozen before any tool runs, so that a claim cannot be quietly replaced by an easier one or similar one. Every tool formulates each claim afresh from the frozen statement, never by translating another tool's code. Otherwise one misreading would be copied into all tools and their agreement would be misleading. Each artefact is read back literally into a sentence (``the code checks the identity for $a=2$, $b=3$ only'') and compared with the claim. For every claim and tool, the stage outputs a decision, a so-called ``verdict'', its strength (general proof, exact certificate, check on examples, numerical evidence) and the back-translation.
\item The fidelity audit compares the artefacts with the claims and records every mismatch as a deviation with a severity (\cref{sec:audit}). A human then reviews the readings chosen for ambiguous claims (H2) and decides each open deviation (H3); deviations that need a retry go back to Stage~0 (red arrow).
\item The record of the collaboration becomes a dialog report that shows which decisions came from whom (\cref{sec:dialog}); the human confirms its borderline items (H4).
\end{enumerate}
Tools are most useful in Stage~1 when finding a result and checking it can be separated. A solver or an LLM may find a certificate, a witness or a proof, and a small trusted checker verifies it \cite{peyrl2008,harrison2007,hales2017}. An error in the untrusted finder can then make a check fail, but it cannot make a false statement pass.

\subsection{Admission criteria for Stage-1 tools}\label{sec:admission}
A tool enters Stage~1 only if its verdicts can be trusted about its artefacts. We require five properties.
\begin{enumerate}[label=(A\arabic*), leftmargin=*, itemsep=1pt]
\item Documented semantics, so that a verdict can be stated as a claim about the artefact.
\item A validation record appropriate to the tool's kind, such as a small trusted kernel, a tested algorithm library, or code verification and benchmark validation \cite{roache1998,oberkampf2010}.
\item Versioning, so that a run can be repeated.
\item Agent access to the complete output, including warnings and failures, through a documented interface such as the Model Context Protocol (MCP) \cite{anthropic2024mcp}.
\item A known ground of trust: formal proof, exact arithmetic, floating-point computation with stated tolerances, or discretisation with error control.
\end{enumerate}
For a tool collection such as ToolUniverse \cite{gao2025tooluniverse}, (A4) holds for all tools at once, but the others must be established tool by tool. \Cref{tab:toolclasses} shows typical domain-specific tools that are certified for use in the given domain. A Stage-1 set should draw on at least two rows and, where possible, include one tool whose verdicts do not depend on tolerances. That tools with different grounds of trust fail in different cases is a design assumption, not an established fact: independently developed programs are known to fail in correlated ways \cite{knight1986}.

\begin{table}[t]
\centering\small
\caption{Classes of validated, agent-usable tools by ground of trust.}
\label{tab:toolclasses}
\begin{tabularx}{\linewidth}{@{}L{2.3cm}L{3.5cm}YY@{}}
\toprule
Class & Examples & A verdict establishes & It does not establish \\
\midrule
Formal proof & Lean~4 + Mathlib \cite{demoura2021lean4,mathlib2020}; Rocq \cite{rocq} & The formal statement follows from the axioms & That the formal statement is the author's claim \\\addlinespace[2pt]
Symbolic computation & Wolfram Language \cite{wolfram} & Identities and decisions under stated assumptions & General results from instance checks; hidden branch choices \\\addlinespace[2pt]
Numerical computation & MATLAB \cite{matlab}, GNU Octave \cite{octave} & Numbers within stated tolerances & Cases near the tolerance; shared formulation errors \\\addlinespace[2pt]
Statistical computation & R \cite{rcore}; SciPy \cite{virtanen2020scipy}, statsmodels \cite{seabold2010statsmodels} & Estimates and tests for the data and model given & That data, model and test are those stated \\\addlinespace[2pt]
Physical simulation & COMSOL \cite{comsol} & Solutions of the discretised model & That model, parameters and mesh are those stated \\
\bottomrule
\end{tabularx}
\end{table}

Even a complementary set does not cover all claims in the same way. Claims with an exact or formal check (identities, inequalities with explicit certificates) are mostly well covered. Claims with strong numerical evidence but no certificate (most results on differential equations, optimality, statistics and simulation) are partly covered. Claims no tool can express, and empirical claims that need validation against data, are not covered.

\section{The fidelity audit}\label{sec:audit}

Bastounis et al.~\cite{bastounis2026} describe the fidelity problem and its limits but give no operational procedure for closing the gap between the user's request as translated by the LLM and what the user wants to do. We turned their analysis into one, packaged as a reusable instruction set for an LLM agent (a \emph{skill}, \texttt{navier-stokes-proving-skill}~\cite{suppl2026}). From the paper we take the two task types, the definition of a faithful translation, the undecidability limit, with ``undecided'' as a legitimate outcome, and the failure patterns behind its examples. The target specification, forbidden acts, deviation protocol, perturbation tests, verdict labels and checklists are our engineering choices; the paper neither prescribes nor endorses them.

The audit starts by writing down, in the author's words, what must be verified: the verbatim claim, its objects, quantifiers, hypotheses, conclusion and presuppositions, whether argument fidelity is required, and which negative outcomes are acceptable. If a field is ambiguous, all readings are listed. The auditor never picks the reading that is easiest to verify. Once this target is frozen, ten acts are forbidden without the human's approval. These are the so-called automated agentic microdecisions that appear in the chain of thought (CoT); however, most human operators do not open and investigate the CoT during operations. They are the concrete forms of ``changing the problem''. Typical microdecision could be replacing the statement by an easier one, adding hypotheses or gap markers without disclosure, silently repairing the author's argument, weakening a test or tolerance, and reporting ``verified'' when only compilation succeeded. Any mismatch between the target and an artefact is a \emph{deviation}, recorded with its location, its type (narrowed scope, weakened test, different statement, unchecked claim, and so on) and what the artefact leaves unestablished. A deviation is \emph{critical} if the statement changed or a success claim would be false. It is \emph{major} if a quantitative detail differs, a claim is left unchecked or a report overstates (or understates) its coverage. It is \emph{minor} if only the representation differs and the equivalence is not yet proved. A critical deviation halts the audit until the human decides (\cref{sec:human}).

The audit itself proceeds in nine steps:
\begin{enumerate}[label=(\arabic*), leftmargin=*, itemsep=0pt]
\item take the frozen target;
\item re-extract the claims from the source;
\item rebuild the \emph{coverage map}, which records for each claim which artefact checks it and how;
\item run the mechanical checks;
\item read each artefact back literally into natural language;
\item run perturbation tests;
\item compare the rebuilt map with the one written in Stage~1;
\item write the deviation log;
\item propose a verdict per claim.
\end{enumerate}
Four of these steps are the most critical ones. The inputs are restricted to the frozen target, the source and the final artefacts, so that the builders' own reasoning is not reused as evidence. Every disagreement between the rebuilt map and the Stage-1 map is a finding to be checked and verified. The mechanical checks, the only tool-dependent part, look for verdicts that are technically correct but say less than they seem: for a proof assistant, the axioms used, gap markers such as \texttt{sorry} and default-value traps such as $\inf\emptyset=0$; for symbolic computation, instance checks reported as general; for numerical computation, tolerances and solver status codes accepted as success; for statistics, the data subset, exclusions and multiple-comparison correction; for simulation, boundary conditions, mesh and units. The literal back-translation gives a signal only, because it must itself be faithful \cite{bastounis2026}. The perturbation tests change one hypothesis, constant or data item at a time and re-run the check and if the check still passes where the target says the item matters, the item was not used.

Each claim receives one of five verdicts:
\begin{itemize}[leftmargin=*, itemsep=0pt]
\item \verdict{FAITHFUL}: the checks found no difference from the claim;
\item \verdict{PROOF-CHANGED}: the claim holds, but by a different argument than the author's;
\item \verdict{STATEMENT-CHANGED}: the artefact establishes something else and answers a different question;
\item \verdict{UNDECIDED}: fidelity cannot be established, for example because the claim is ambiguous or an equivalence is unproven;
\item \verdict{NOT PROVEN}: a faithful attempt failed, and the gaps are listed.
\end{itemize}
When several apply, the worst wins, in the order \verdict{STATEMENT-CHANGED}, \verdict{UNDECIDED}, \verdict{NOT PROVEN}, \verdict{PROOF-CHANGED}, \verdict{FAITHFUL}; a changed statement comes first because a reader would take it as an answer to the author's question.

\section{Importance of human verification}\label{sec:human}

Some fidelity questions have no algorithmic answer (\cref{sec:problem}); others have one only after somebody chooses between readings: whether ``natural numbers'' include $0$, whether a printed decimal is exact or rounded, whether ``the largest value'' is an attained maximum. The choice belongs to the person responsible for the claim. The framework therefore keeps a human in the loop, but only at four decision points (H1--H4 in \cref{fig:pipeline}), so that human attention goes to what only the person in charge of the research can decide.

The use-case we design for is checking many AI-produced results, possibly unattended. The two example toolchains of \cref{sec:toolchains} therefore run fully automatically and ask nothing once started. This moves the human decisions to the boundaries of a run rather than removing them, and five strict rules make that faithful.
\begin{enumerate}[label=\arabic*., leftmargin=*, itemsep=1pt]
\item A reading the human states in the input is used as stated.
\item Every other reading is fixed by a published default rule before any tool runs, and recorded with the rule's name. Since the rule never depends on a result, the run cannot pick the reading that is easiest to confirm.
\item A material ambiguity, one where one reading could be true while the other is false, is not guessed but split: both readings run as separate claims.
\item Words follow evidence: ``proved'' only for a general proof or an exact certificate, ``checked on examples'' for an instance check, ``numerically supported'' for floating-point evidence. A statistical result is never called a proof.
\item If a tool cannot be reached, the run aborts and says why instead of switching to another route.
\end{enumerate}

The human decides at four points.
\begin{enumerate}[label=H\arabic*., leftmargin=*, itemsep=1pt]
\item Before the run, the human states the claims, the acceptance conditions and any reading that must not be left to a default rule.
\item After Stage~1, the human reviews the readings the run chose, the split claims and every back-translation that is only a partial match; if a reading is not what was meant, the claim is restated and a new run started, and the old run stays on record.
\item After the audit, the human decides each open deviation: retry faithfully, accept it with a label, amend the target as a new, explicitly approved version, or stop with a negative verdict that lists the gaps.
\item At the end, the human accepts or rejects the verdicts, decides what to change in the paper, and confirms the coding of borderline items in the dialog report.
\end{enumerate}

At scale, a human cannot read every artefact within a reasonable time, but the records can direct attention. Claims with a split reading, a partial back-translation, disagreeing tools, an inconclusive or refuted result, or a critical deviation need a human verification or decision. Claims proved by tools on different grounds of trust with exact back-translations need less. This is a proposal for triage, not a guarantee: a claim misread the same way by every tool will look clean, and only a human who knows what was meant will notice. The framework does not make that check unnecessary; instead, it provides the record against which it can be done.

\section{The human--AI dialog report}\label{sec:dialog}

The human cannot be removed from forming and checking results, and, to avoid misleading readers, the human's share must be visible. Readers need to know which decisions a person took responsibility for, and some journals require a statement of AI use \cite{nature2023rules}. Moreover, the record of a collaboration shows errors that no output check reveals, for example an AI that ran experiments before their design had been agreed. Contributor role taxonomies such as the Contributor Roles Taxonomy (CRediT) \cite{brand2015credit} describe roles, but give no method to reconstruct from the record which decisions came from whom.

The dialog-report method, developed by the third author (R.~Holl\'os) and implemented as the skill \texttt{ai-human-dialog-report}~\cite{suppl2026}, turns the record of a finished collaboration into a short report about how the result evolved. Before writing, the human fixes the audience, the attribution style, the style register of the report (direct, naming the AI's mistakes plainly, or diplomatic) and the privacy level (for example, whether theoretical ideas that the authors shared in a brainstorming session may appear in the report). These choices shape the framing, but the contribution facts remain. Only turns that carry content are numbered, and the human is quoted verbatim, in the original language, so that the report can be checked against the record. The central count is the \emph{correction ledger}: the turns that changed the next technical move, the scope or an acceptance condition. They are counted in both directions, from human to AI and from AI to human. Changes of language or format are counted separately, and zero is a legitimate count. Turns after which the next answer would have been different are singled out as the so-called \emph{inflection points}. An inflection point is a point in the research process at which the direction of the research changes, either because the human decides to change it or because of a microdecision of the LLM. Each comes with the decisive quotation, a two-column attribution (human contribution, AI contribution) and what the AI could have done better. The method deliberately gives no overall ``AI share'': a conceptual correction, thousands of generated lines of code and the author's final acceptance are not commensurable. Because counts depend on the coding rule, the report states the rule and gives both counts for borderline items. The counts describe the record, they do not measure authorship.

The audit compares artefacts with claims; the dialog report compares turns with decisions. They meet at the human: a deviation decided at H3 is an inflection point for the dialog report, and a reading fixed at H1 or H2 is a human contribution that the report records.

\section{Two examples built with the proposed methodology}\label{sec:toolchains}

The framework of \cref{sec:pipeline,sec:audit,sec:human,sec:dialog} does not name any tool. To use it in a particular field, one builds a concrete tool layer by making five design choices, each of which follows from an earlier section.
\begin{enumerate}[label=\arabic*., leftmargin=*, itemsep=1pt]
\item Choose the Stage-1 tools: tools that meet the admission criteria A1--A5 and rest on at least two different grounds of trust (\cref{tab:toolclasses}).
\item Fix how the tools work together: the claims are frozen first, every tool formulates them independently, and the tools run one after another.
\item Write default reading rules for the ambiguities typical of the field, and a rule for when an ambiguity is material and must be split (\cref{sec:human}).
\item Define a vocabulary of evidence that matches the grounds of trust, and a rule for comparing the tools' results.
\item Define the record that the audit and the human will use: literal back-translations, timestamps, an abort rule, a validated result file and a readable report.
\end{enumerate}
We give two examples of this construction, one for mathematics and one for statistics. Both are packaged as agent skills, that is, written instructions with embedded scripts that an LLM agent follows step by step~\cite{suppl2026}, and both call their tools through MCP servers \cite{anthropic2024mcp} on the researcher's computer. Both implement Stage~1 together with the literal back-translation. The audit, the human decisions and the dialog report remain separate steps. \Cref{tab:toolchains} sets the two examples side by side, choice by choice.

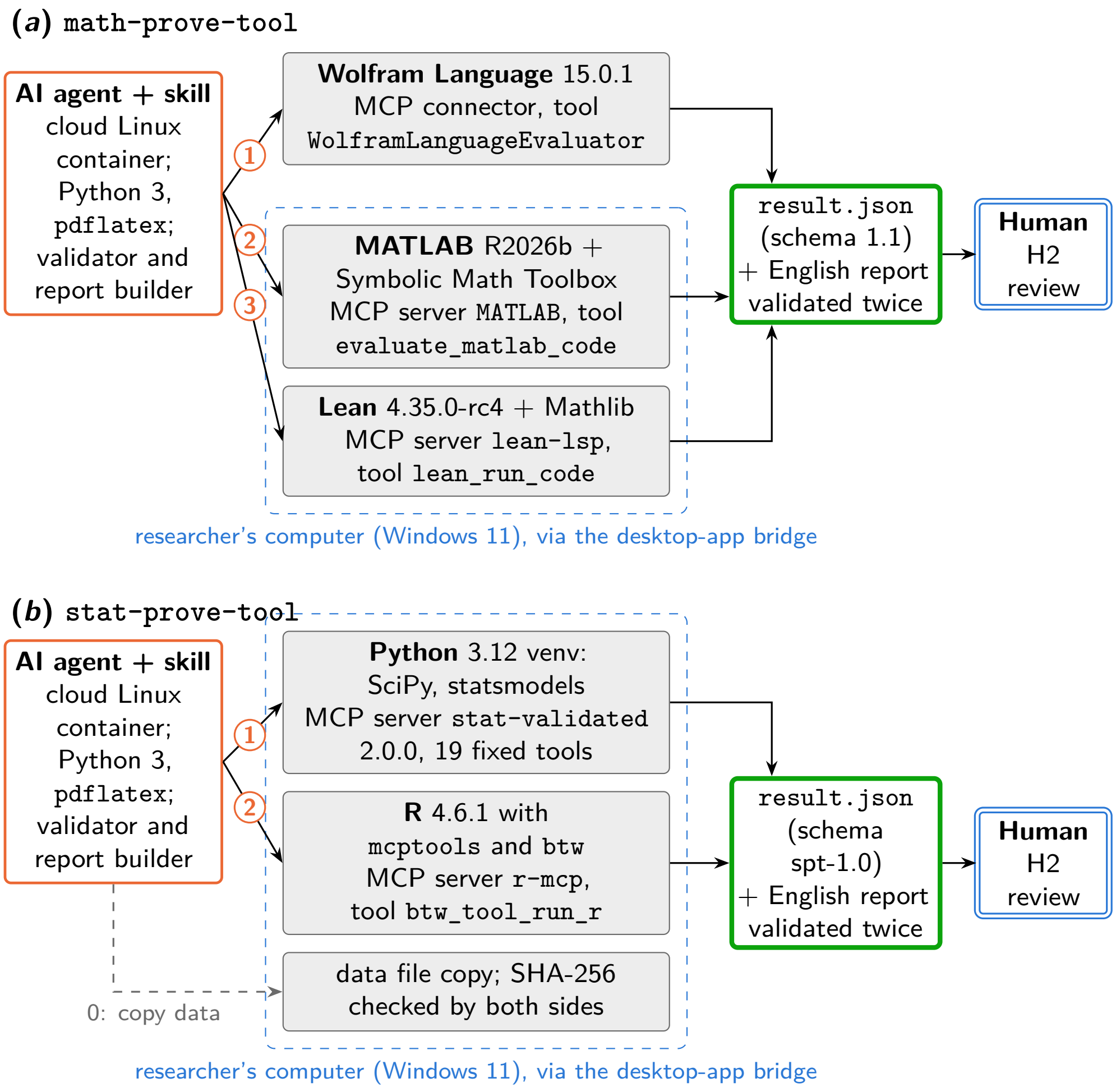
\begin{figure}[t]
\centering
\begin{tikzpicture}[
  node distance=4mm and 6mm,
  box/.style={draw, rounded corners=2pt, align=center, font=\footnotesize, fill=white, minimum height=11mm},
  ai/.style={box, draw=corange, line width=0.9pt, text width=23mm},
  tool/.style={box, draw=cgray, fill=cgray!12, text width=43mm},
  outbox/.style={box, draw=cgood, line width=1.6pt, text width=22mm},
  human/.style={box, draw=cblue, line width=0.6pt, double, double distance=1pt, text width=13mm},
  num/.style={circle, draw=corange, fill=white, text=corange, font=\scriptsize\bfseries, inner sep=0.6pt, minimum size=3.6mm},
  arr/.style={-{Stealth[length=1.8mm]}, line width=0.6pt},
  lab/.style={font=\small\bfseries, anchor=west}
]
\node[lab] at (-1.35,2.0) {(\textit{a}) \texttt{math-prove-tool}};
\node[ai] (ag) at (0,0) {\textbf{AI agent + skill}\\cloud Linux container;\\Python~3, \texttt{pdflatex};\\validator and\\report builder};
\node[tool, right=7mm of ag, yshift=10mm] (wl) {\textbf{Wolfram Language} 15.0.1\\MCP connector, tool \texttt{WolframLanguageEvaluator}};
\node[tool, below=7mm of wl] (ml) {\textbf{MATLAB} R2026b + Symbolic Math Toolbox\\MCP server \texttt{MATLAB}, tool \texttt{evaluate\_matlab\_code}};
\node[tool, below=2mm of ml] (ln) {\textbf{Lean} 4.35.0-rc4 + Mathlib\\MCP server \texttt{lean-lsp}, tool \texttt{lean\_run\_code}};
\begin{scope}[on background layer]
\node[draw=cblue, dashed, rounded corners=3pt, fit=(ml)(ln), inner sep=2mm, label={[font=\scriptsize, text=cblue, align=center]below:researcher's computer (Windows~11), via the desktop-app bridge}] (pc1) {};
\end{scope}
\node[outbox, right=7mm of ml, yshift=5mm] (o1) {\texttt{result.json} (schema 1.1)\\+ English report\\validated twice};
\node[human, right=4mm of o1] (h1) {\textbf{Human}\\H2 review};
\draw[arr] (ag.east) -- node[num, pos=0.45]{1} (wl.west);
\draw[arr] (ag.east) -- node[num, pos=0.45]{2} (ml.west);
\draw[arr] (ag.east) -- node[num, pos=0.45]{3} (ln.west);
\draw[arr] (wl.east) -| ([xshift=5mm]o1.north west);
\draw[arr] (ml.east) -- (ml.east -| o1.west);
\draw[arr] (ln.east) -| ([xshift=5mm]o1.south west);
\draw[arr] (o1) -- (h1);
\begin{scope}[yshift=-67mm]
\node[lab] at (-1.35,1.75) {(\textit{b}) \texttt{stat-prove-tool}};
\node[ai] (ag2) at (0,0) {\textbf{AI agent + skill}\\cloud Linux container;\\Python~3, \texttt{pdflatex};\\validator and\\report builder};
\node[tool, right=7mm of ag2, yshift=7mm] (py) {\textbf{Python} 3.12 venv: SciPy, statsmodels\\MCP server \texttt{stat-validated} 2.0.0, 19 fixed tools};
\node[tool, below=2mm of py] (rr) {\textbf{R} 4.6.1 with \texttt{mcptools} and \texttt{btw}\\MCP server \texttt{r-mcp}, tool \texttt{btw\_tool\_run\_r}};
\node[tool, below=2mm of rr, minimum height=7mm] (dd) {data file copy; SHA-256 checked by both sides};
\begin{scope}[on background layer]
\node[draw=cblue, dashed, rounded corners=3pt, fit=(py)(rr)(dd), inner sep=2mm, label={[font=\scriptsize, text=cblue]below:researcher's computer (Windows~11), via the desktop-app bridge}] (pc2) {};
\end{scope}
\node[outbox, right=7mm of rr] (o2) {\texttt{result.json} (schema spt-1.0)\\+ English report\\validated twice};
\node[human, right=4mm of o2] (h2) {\textbf{Human}\\H2 review};
\draw[arr] (ag2.east) -- node[num, pos=0.45]{1} (py.west);
\draw[arr] (ag2.east) -- node[num, pos=0.45]{2} (rr.west);
\draw[arr, draw=cgray, dashed] (ag2.south) |- node[pos=0.62, below, font=\scriptsize, text=cgray]{0: copy data} (dd.west);
\draw[arr] (py.east) -| ([xshift=5mm]o2.north west);
\draw[arr] (rr.east) -- (o2.west);
\draw[arr] (o2) -- (h2);
\end{scope}
\end{tikzpicture}
\caption{Wiring of the two examples. Single frames (orange): the AI agent with the skill and its scripts; shaded frames (grey): Stage-1 tools behind MCP servers, with the version used in this paper; numbers: the fixed order of the tool runs; dashed outline (blue): tools that run on the researcher's computer; heavy frames (green): the validated record that goes to the human (double frames, decision point H2).}
\label{fig:wiring}
\end{figure}

Choices 2 and 5 are realised in the same way in both examples. The input is never modified, and errors in it are reported, not fixed. The claims are extracted verbatim and frozen as a versioned list before any tool runs. Each tool writes its own formulation from that list, never from another tool's code or output. Every call is timestamped from the system clock and stays on record, failed calls included. A pre-flight check confirms that every server answers; a tool that still fails after its retries aborts the run, and workarounds through another route are forbidden. Finally, an embedded validator checks the result file against a fixed schema (timestamps, order of tools, file hashes, consistency of verdicts and counts) before and after a report is generated.

\subsection{Example 1: mathematics (\texttt{math-prove-tool})}
The claims are theorems, identities and numerical statements in any document format. For choice~1, the example uses three tools with different grounds of trust: the Wolfram Language (symbolic decision procedures), MATLAB (symbolic and numerical computation) and Lean~4 with Mathlib (formal proof), run in this order on every claim. For choice~3, ten default rules (D1--D10) fix the usual ambiguities of mathematical text. For example, $\mathbb N$ contains $0$, as in Mathlib; free variables are universally quantified; and a rounded number such as $0.73$ stands for the interval of its precision, against which an exactly computed value is tested. For choice~4, each result is classified by the kind of evidence it provides: general proof; exact certificate, that is, a concrete object checked in exact arithmetic; symbolic instance; numerical evidence with a stated tolerance; failed; or not expressible. Its outcome is confirmed, refuted (only with a counterexample or a proof of the negation), inconclusive or error, and the comparison reports the strongest confirmed evidence per claim and whether the tools agree. A Lean result counts as a proof only if \texttt{\#print axioms} lists nothing beyond the three standard axioms and no \texttt{sorry} occurs. Where a claim is a numerical matrix inequality, the example applies the untrusted-finder, trusted-checker principle of Stage~1: a floating-point solver proposes a certificate, which is rounded to rationals and factorised exactly, and Lean checks the factorisation with the paper's expression in the theorem statement.

\subsection{Example 2: statistics (\texttt{stat-prove-tool})}
The input is one or more data files and an analysis plan, optionally with claims such as ``group B scores higher than group A''. For choice~1, the example uses two independently implemented statistical environments, run in this order. The first is a local MCP server, \texttt{stat-validated} (version 2.0.0 or later), which exposes a fixed catalogue of tested procedures built on SciPy \cite{virtanen2020scipy}, statsmodels \cite{seabold2010statsmodels} and related Python libraries: descriptive statistics, $t$- and rank tests, analysis of variance (ANOVA), $\chi^2$ and Fisher tests, correlations, linear and generalised linear models, and principal components. A method outside the catalogue is reported as not available and never imitated with other tools. The second is base R with its installed packages \cite{rcore}, reached through the \texttt{mcptools} and \texttt{btw} packages \cite{mcptools,btw}; the skill installs nothing. The pre-flight confirms that both environments see the same data file (same hash, rows and columns). For choice~2, all code of both environments is written and frozen before the first call, so that no setting can be adjusted after a result is seen. For choice~3, fourteen default rules (S1--S14) fix open analysis details, for example $\alpha=0.05$, two-sided tests, Welch's test when variances are not stated, and no multiplicity correction unless stated. A comparison whose test family is not named is split into a parametric and a rank-based variant, and a directional claim without a stated test direction into a two-sided and a one-sided variant. For choice~4, a claim is decided by a rule fixed in advance (for example $p<\alpha$ and a positive effect) as supported, not supported or inconclusive, and the report never calls a claim proved, true or false. The two environments are compared number by number, with relative tolerances of $10^{-6}$ for closed-form quantities and $10^{-4}$ for iterative fits.

\subsection{Environment and wiring}\label{sec:wiring}
\Cref{fig:wiring} shows how the two examples are connected, and \cref{tab:env} lists the software versions used in this paper. The agent (Claude, model \texttt{claude-opus-5-5}) works in a Linux cloud container. There it writes the formulation files, takes the timestamps, keeps the run directory and runs the two embedded scripts: the validator and the report builder, both standard-library Python~3, the latter calling \texttt{pdflatex}. The tools are reached as MCP servers. The Wolfram Language is reached through an MCP connector whose kernel reported version 15.0.1 for Linux x86-64. All other tools run on the researcher's Windows~11 computer and are reached through the bridge of the Claude desktop application, which exposes the computer's local MCP servers to the agent.

One tool call has the same shape in both examples. The agent saves the formulation as a file, hashes it, records the start time, sends the code (or, for \texttt{stat-validated}, a tool name with arguments in JavaScript Object Notation, JSON) through the MCP server, records the end time and saves the complete raw output next to the formulation. In \texttt{math-prove-tool}, each Wolfram call is self-contained, MATLAB code writes its output to a \texttt{diary} log that is read back, and Lean code is compiled as a snippet that imports Mathlib, with \texttt{\#print axioms} for every theorem; several Lean claims may share one module so that Mathlib loads only once. The Lean server often needs a cold start, so its pre-flight is retried up to five times with growing waits, and a Lean build is never started from the skill. In \texttt{stat-prove-tool}, the data file is first copied into a connected folder on the computer under a run-specific path, and both sides must report the same SHA-256 hash, row count and column names before any task runs. Python tasks call one of the 19 fixed tools of \texttt{stat-validated}; R tasks are self-contained scripts for \texttt{btw\_tool\_run\_r} that return their results through \texttt{jsonlite}. The skill installs no packages and changes no setting on the computer.

\begin{table}[t]
\centering\small
\caption{The two examples, design choice by design choice.}
\label{tab:toolchains}
\begin{tabularx}{\linewidth}{@{}L{3.1cm}YY@{}}
\toprule
Design choice & Example 1: \texttt{math-prove-tool} & Example 2: \texttt{stat-prove-tool} \\
\midrule
1. Tools and grounds of trust & Wolfram Language (symbolic decision), MATLAB (symbolic and numerical), Lean~4 + Mathlib (formal proof) & Python \texttt{stat-validated} server and R: two independent statistical implementations \\\addlinespace[2pt]
2. Working together & Frozen claims; independent formulation per tool; Wolfram $\to$ MATLAB $\to$ Lean & Frozen tasks; all code of both sides frozen before the first call; Python $\to$ R \\\addlinespace[2pt]
3. Default readings and splits & D1--D10; split of one material ambiguity per claim & S1--S14; split by test family or test direction \\\addlinespace[2pt]
4. Evidence and comparison & Evidence type and outcome; strongest evidence; agree, partial or disagree & Supported, not supported or inconclusive; number-by-number agreement within tolerance tiers \\\addlinespace[2pt]
5. Record & Validated JSON (schema 1.1) and an English report & Validated JSON (schema spt-1.0) and an English report \\
\bottomrule
\end{tabularx}
\end{table}

\begin{table}[t]
\centering\small
\caption{Software environment of the two examples, as reported by the tools' pre-flight checks.}
\label{tab:env}
\begin{tabularx}{\linewidth}{@{}L{3.0cm}L{3.6cm}Y@{}}
\toprule
Component & Where it runs & Version and connection \\
\midrule
AI agent & Cloud Linux container & Claude, \texttt{claude-opus-5-5}; Python~3 (standard library) and TeX Live \texttt{pdflatex} for the scripts \\\addlinespace[2pt]
Bridge & Researcher's computer & Claude desktop application 3.1623.0, Windows~11 (build 26300) \\\addlinespace[2pt]
Wolfram Language & MCP connector & Kernel 15.0.1 for Linux x86-64; tool \texttt{WolframLanguageEvaluator} \\\addlinespace[2pt]
MATLAB & Researcher's computer & R2026b (26.2) with Symbolic Math Toolbox 26.2; MCP server \texttt{MATLAB} \\\addlinespace[2pt]
Lean & Researcher's computer & Lean 4.35.0-rc4 with Mathlib; MCP server \texttt{lean-lsp} \\\addlinespace[2pt]
Python statistics & Researcher's computer & \texttt{stat-validated} 2.0.0; Python 3.12.10, NumPy 2.5.4, SciPy 1.18.1, statsmodels 0.15.0, pandas 3.0.6, scikit-learn 1.9.1 \\\addlinespace[2pt]
R statistics & Researcher's computer & R 4.6.1; \texttt{mcptools} 1.0.3, \texttt{btw} 1.5.0, \texttt{jsonlite} 2.0.0; MCP server \texttt{r-mcp} \\
\bottomrule
\end{tabularx}
\end{table}
Neither example decides fidelity; both make the material for that decision explicit: the frozen list, the readings and the rules that chose them, and a literal back-translation of every artefact. In statistics, two environments that agree establish that the computation is right for those data and that plan, not that the plan is the right analysis. The same five choices can be made for other fields, for example with a simulator such as COMSOL and an independent solver for physical models; we have not built such an instance.

\subsection{Use of AI in this study}\label{sec:aiuse}
This paper is about AI-assisted research, and AI tools were used throughout; LLMs are not authors of this work. The skills were written with the help of an AI assistant (Claude, Anthropic) under the direction of the first author. The example run of \cref{sec:example} was executed by the same assistant (configured as \texttt{claude-opus-5-5}) through tool connections to a Wolfram kernel and to MATLAB and Lean installations on the first author's computer (\cref{sec:wiring}). The manuscript was drafted by the assistant from an earlier version revised by the second author, from the skill files and from the run record, following a plan approved by the first author, and was then reviewed and revised by the authors. All numbers in \cref{sec:example} were taken from the run record. The authors take full responsibility for the content of the paper.

\section{A small working example}\label{sec:example}

We ran \texttt{math-prove-tool} on two short claims, each with one typical ambiguity. The input file contained, verbatim:
\begin{quote}\small
P1. For every natural number $n$, $6$ divides $n^3-n$.\\
P2. The largest value of $f(x)=x\,e^{-x}$ on $[0,\infty)$ is $0.37$.
\end{quote}

The experiment was done on 21 October 2026. All three servers answered on the first attempt: Wolfram Language 15.0.1, MATLAB R2026b with the Symbolic Math Toolbox, and Lean 4.35.0-rc4 with Mathlib. The skill then found three ambiguities and resolved them before any tool ran (\cref{tab:readings}). For P1, whether $\mathbb N$ contains $0$ does not matter, since $6$ divides $0$; $n^3-n$ is read in the integers. For P2, the decimal $0.37$ matters, since the exact reading could be false while the rounded one is true, so the skill split it into P02a (the maximum is exactly $37/100$) and P02b (the maximum lies in $[0.365, 0.375)$).

\begin{table}[!htb]
\centering\small
\caption{Readings fixed by the skill before any tool ran.}
\label{tab:readings}
\begin{tabularx}{\linewidth}{@{}L{1.3cm}L{4.4cm}YL{2.1cm}@{}}
\toprule
Claim & Ambiguity & Reading used & Chosen by \\
\midrule
P01 & Does $\mathbb N$ contain $0$? & Yes; not material, since $n=0$ gives $0$ & Default rule D1 \\\addlinespace[2pt]
P01 & $n^3-n$ in $\mathbb N$ or in $\mathbb Z$? & Ordinary subtraction in $\mathbb Z$ & Default rule D10 \\\addlinespace[2pt]
P02a/b & Is $0.37$ exact or rounded? Material & P02a: $M=37/100$; P02b: $M\in[0.365,0.375)$ & Both readings run (split) \\
\bottomrule
\end{tabularx}
\end{table}

Each tool wrote its own code from the frozen statements. \Cref{tab:example} shows the results.

\begin{table}[!htb]
\centering\small
\caption{Results of the example run: evidence type and outcome per claim and tool.}
\label{tab:example}
\begin{tabularx}{\linewidth}{@{}L{1.1cm}YYYL{1.75cm}@{}}
\toprule
Claim & Wolfram Language & MATLAB & Lean & Agreement \\
\midrule
P01 & general proof, confirmed & general proof, confirmed (back-translation partial) & general proof, confirmed & agree \\\addlinespace[2pt]
P02a & general proof, \textbf{refuted} & symbolic instance, inconclusive & general proof, \textbf{refuted} & partial \\\addlinespace[2pt]
P02b & general proof, confirmed & symbolic instance, inconclusive & general proof, confirmed & partial \\
\bottomrule
\end{tabularx}
\end{table}

The Wolfram Language run took 39~s for the three claims. For P1, \texttt{FullSimplify} returned \texttt{True} for $n^3-n\equiv 0 \pmod 6$ over the non-negative integers, and \texttt{Reduce} proved that no counterexample exists. For P2, \texttt{MaxValue} gave the exact maximum $1/e$, attained at $x=1$, and \texttt{Resolve} confirmed $f(x)\le 1/e$ for all $x\ge 0$. Hence $M=1/e=0.36788\ldots$, which is not $37/100$ (P02a refuted) but lies in $[0.365,0.375)$ (P02b confirmed). Internal messages (\texttt{Infinity::indet}) during the maximisation were recorded as warnings.

The MATLAB run took 33~s. For P1, the direct test \texttt{isAlways(mod(n\^{}3-n,6)==0)} returned ``Unable to prove''. The code then wrote $n=6k+r$ with a symbolic integer $k$ and checked, for $r=0,\dots,5$, that every coefficient of $(6k+r)^3-(6k+r)$ is divisible by $6$. This covers every $n$ and was classified as a general proof, but the back-translation is marked \emph{partial}: the reduction to six remainders is stated in a code comment, not checked by MATLAB. For P2, MATLAB found the only critical point $x=1$ and the exact value $e^{-1}$, but \texttt{isAlways} could not prove $f(x)\le e^{-1}$ for $x\ge 0$. It therefore showed what the value at the critical point is, not that this value is the maximum, and both readings stayed inconclusive.

The Lean run took 19~s, with one module for all three claims. The theorems state the claims directly. For example, the exact reading of P2 was formalised as its negation,
\begin{center}\small
\texttt{theorem P02a :~}$\neg$\,\texttt{IsGreatest ((fun x :~}$\mathbb{R}$\texttt{~=> x * Real.exp (-x)) '{}' Set.Ici 0) (37 / 100)}
\end{center}
and proved from $x e^{-x}\le e^{-1}$ (a consequence of $t+1\le e^{t}$) and Mathlib's certified bound $e>2.7182818283$. P1 was proved by induction, and P02b by exhibiting $M=e^{-1}$ with the bounds $0.365\le e^{-1}<0.375$. All three theorems depend only on the three standard axioms, and no \texttt{sorry} occurs.

Small as it is, the run exhibits four effects the framework is designed for.
\begin{enumerate}[label=\arabic*., leftmargin=*, itemsep=1pt]
\item The ambiguous claim was not given the reading that is easiest to confirm. It was split, and the two readings received opposite results between which a human must choose.
\item The tools reached different strengths on the same claims, and the cross-tool table shows this rather than averaging it away.
\item MATLAB's proof of P1 rested on a step outside the tool, and the literal back-translation flagged exactly that step.
\item The report kept the strengths apart in its wording: ``proved in general'' for the Wolfram and Lean results, ``instance check'' and ``inconclusive'' for MATLAB's result on P2.
\end{enumerate}
The validator accepted the result file before and after the report was built.

The run ends Stage~1. At H2 the human would look at \cref{tab:readings}: if the author meant the rounded value, P02b is the relevant result and is proved; if the exact value, the claim is false and the paper should print $1/e$. No tool can make this choice. The audit would then compare each artefact with the claim as decided; for instance, it would check that the Lean statement contains the function and interval of the claim rather than a precomputed number. The dialog report would record that the AI proposed the two claims, that the human accepted them, and that the reading of P2 is the human's choice. We did not run the audit or the dialog report on this example.

\section{Discussion and limitations}\label{sec:discussion}

Claims, artefacts and verdicts exist wherever tools check results, and so do the two gaps of \cref{fig:gap}; what changes between domains is the tool set, the mechanical checklists and the default rules, as the two examples show. In statistics and simulation fewer claims admit an exact check, and comparing the artefact's inputs with the paper's description will matter more. Domain independence is a property of the method's structure; we have not tested it empirically outside mathematics, and the statistical toolchain is described here, not evaluated.

The framework and its present implementation have four limitations.
\begin{enumerate}[label=\arabic*., leftmargin=*, itemsep=1pt]
\item Running the toolchains without questions is what makes checking at scale possible, but it has a price: a default reading can differ from the author's intention. The framework does not hide this. Every automatic choice is recorded with its rule, material ambiguities are run in both readings, and the human reviews the choices at H2. A reading that is wrong but not material, and that every tool adopts, will still pass; only a human who knows what was meant can catch it.
\item In our implementation the formulations for all tools, the back-translations and the audit are written by the same AI. Independent formulation from the frozen claims prevents one tool's code from being copied into another, but it does not prevent the same misreading from being made three times. An auditor that shares the builder's training may share its blind spots. Repeating the audit on a sample with a different model or a human reviewer is the obvious remedy. The same holds for the design assumption that tools with different grounds of trust fail in different cases \cite{knight1986}.
\item The example of \cref{sec:example} illustrates the mechanics on two claims; it is not an evaluation. We have not measured how often the framework finds deviations, how much human time it saves or costs, or what a careful human reader would find without it. Known-answer test sets, in which the correct verdict for each claim is fixed before the run, are the natural next step.
\item The framework checks whether artefacts match claims and whether tools compute correctly; whether a model describes the real system remains a question of validation against data \cite{oberkampf2010}. The paper that motivates the audit \cite{bastounis2026} is a preprint, and we rely only on its examples and its stated theorem. The account of the 2026 manuscript release rests on the release's own repository and a news report \cite{openai2026math,nature2026openai}.
\end{enumerate}

\section{Conclusion}\label{sec:conclusion}

AI systems now produce research results faster than people can referee them, so most of these results will be checked by software that AI agents call. A passing check certifies the artefact that was checked, not the claim the author made, and no procedure can decide in general whether the two match. The human therefore stays in the loop, and once the human stays, the human's share of the work must be recorded.

The proposed framework meets these requirements with four parts: validated, agent-usable tools, a fidelity audit, human verification at four decision points, and a dialog report. Two examples built with the methodology, automatic toolchains for mathematics and statistics, show how the tool layer is instantiated: they freeze the claims, let every tool formulate independently, split material ambiguities instead of guessing, match their words to the evidence and validate their own records. A small run showed these mechanics at work: one reading of an ambiguous claim was refuted, the other proved, and one tool stayed inconclusive where two proved.

We recommend that researchers who use AI agents admit tools by explicit criteria and combine tools with different grounds of trust. They should keep the frozen claims and the chosen readings with the results, and record who decided what. Journals receiving AI-assisted work could ask for these records together with the AI-use statement. For each claim, the records show whether it rests on a formal proof, on a computation or on no check, and which choices a human made.

\section*{Data availability}
The skill files \texttt{math-prove-tool}, \texttt{stat-prove-tool}, \texttt{navier-stokes-proving-skill} and \texttt{ai-human-dialog-report}, and the complete run directory of the example in \cref{sec:example} (input, frozen claim list, formulation files, raw tool outputs, event ledger, \texttt{result.json}, the generated report and the two scripts as used), with a README and a SHA-256 manifest, are archived on Zenodo~\cite{suppl2026}.

\section*{Declaration of AI use}
AI tools were used in this study and in preparing the manuscript, as described in \cref{sec:aiuse}. In short: the skills were written with the help of an AI assistant (Claude, Anthropic) under the direction of the first author; the example run was executed by the same assistant (\texttt{claude-opus-5-5}) through the tool connections of \cref{sec:wiring}; and the manuscript was drafted by the assistant from an earlier version revised by the second author, from the skill files and from the run record, following a plan approved by the first author, and was then reviewed and revised by the authors. All numbers in \cref{sec:example} were taken from the run record. AI tools are not authors of this work, and the authors take full responsibility for its content.

\section*{Competing interests}
The authors declare no competing interests.

\section*{Funding}
Gy\"orgy Eigner was supported by the Distinguished Program of Obuda University.

\section*{Acknowledgements}
We gratefully acknowledge the support of the HUN-REN AI Service Center, which provides critical resources and infrastructure in the form of inspiration, education, consultation, and technology to the research community. Its mission to address researchers' needs in AI and to enable world-class results in scientific inquiry has been instrumental in advancing our work.

{\small
\bibliographystyle{RS}
\bibliography{refs}}

\end{document}